\documentclass[lettersize,journal]{IEEEtran}
\usepackage{amsmath,amsfonts}
\usepackage{algorithm}
\usepackage{array}
\usepackage[caption=false,font=small,labelfont=sf,textfont=sf]{subfig}
\usepackage{textcomp}
\usepackage{stfloats}
\usepackage{url}
\usepackage{verbatim}
\usepackage{graphicx}
\usepackage{cite}
\usepackage{color}
\usepackage{algpseudocode}
\usepackage{siunitx}
\usepackage{bm}
\usepackage{booktabs} 
\usepackage{makecell}

\definecolor{blue_tab}{rgb}{0.12156863, 0.46666667, 0.70588235}
\definecolor{red_tab}{rgb}{0.83921569, 0.15294118, 0.15686275}
\definecolor{hpp_tab}{rgb}{0.6196078431372549, 0.792156862745098, 0.8823529411764706}
\definecolor{hpp_opt_tab}{rgb}{0.9921568627450981, 0.6823529411764706, 0.4196078431372549}
\definecolor{pddl_tab}{rgb}{0.6196078431372549, 0.6039215686274509, 0.7843137254901961}
\definecolor{our_tab}{rgb}{0.984313725490196, 0.5019607843137255, 0.4470588235294118}
\definecolor{our_opt_tab}{rgb}{0.6313725490196078, 0.8509803921568627, 0.6078431372549019}
\definecolor{blue}{rgb}{0,0,1}

\def\eg{\emph{e.g.}} 
\def\ie{\emph{i.e.}}

\def\wrt{w.r.t.} 

\makeatother

\begin{document}

\title{Multi-step manipulation task and motion planning guided by video demonstration}

\author{Kateryna Zorina$^{\clubsuit}$ \quad David Kovar$^{\clubsuit}$ \quad Mederic Fourmy$^{\clubsuit}$ \quad  Florent Lamiraux$^{\diamondsuit}$ \\
Nicolas Mansard$^{\diamondsuit}$ \quad Justin Carpentier$^\spadesuit$ \quad  Josef Sivic$^{\clubsuit}$ \quad  Vladimir Petrik$^{\clubsuit}$
\thanks{$^{\clubsuit}$ CIIRC, Czech Technical University in Prague}%
\thanks{$^{\diamondsuit}$ LAAS-CNRS, Universite de Toulouse, CNRS, Toulouse}%
\thanks{$^{\spadesuit}$ INRIA, Paris}%
\thanks{This work is part of the AGIMUS project, funded by the European Union under GA no.101070165. Views and opinions expressed are, however, those of the author(s) only and do not necessarily reflect those of the European Union or the European Commission. Neither the European Union nor the European Commission can be held responsible for them. This work was funded by the European Regional Development Fund under the project IMPACT (reg. No. CZ.02.1.01/0.0/0.0/15\_003/0000468), the Grant Agency of the CTU in Prague, grant No. SGS21/178/OHK3/3T/17, the Louis Vuitton ENS Chair on Artificial Intelligence, and by the European Union's Horizon Europe projects euROBIN (No. 101070596). }
}



\maketitle
\begin{abstract}
This work aims to leverage instructional video to solve complex multi-step task-and-motion planning tasks in robotics.
Towards this goal, we propose an extension of the well-established Rapidly-Exploring Random Tree (RRT) planner, which simultaneously grows multiple trees around grasp and release states extracted from the guiding video.
Our key novelty lies in combining contact states and 3D object poses extracted from the guiding video with a traditional planning algorithm that allows us to solve tasks with sequential dependencies, for example, if an object needs to be placed at a specific location to be grasped later.
We also investigate the generalization capabilities of our approach to go beyond the scene depicted in the instructional video.
To demonstrate the benefits of the proposed video-guided planning approach, we design a new benchmark with three challenging tasks: (i)~3D re-arrangement of multiple objects between a table and a \textit{shelf}, (ii)~multi-step transfer of an object through a \textit{tunnel}, and (iii)~transferring objects using a tray similar to a \textit{waiter} transfers dishes.
We demonstrate the effectiveness of our planning algorithm on several robots, including the \textit{Franka Emika Panda} and the \textit{KUKA KMR iiwa}. For a seamless transfer of the obtained plans to the real robot, we develop a trajectory refinement approach formulated as an optimal control problem~(OCP).
\end{abstract}

\begin{figure}[t]
  \centering

    \includegraphics[width=0.48\textwidth]{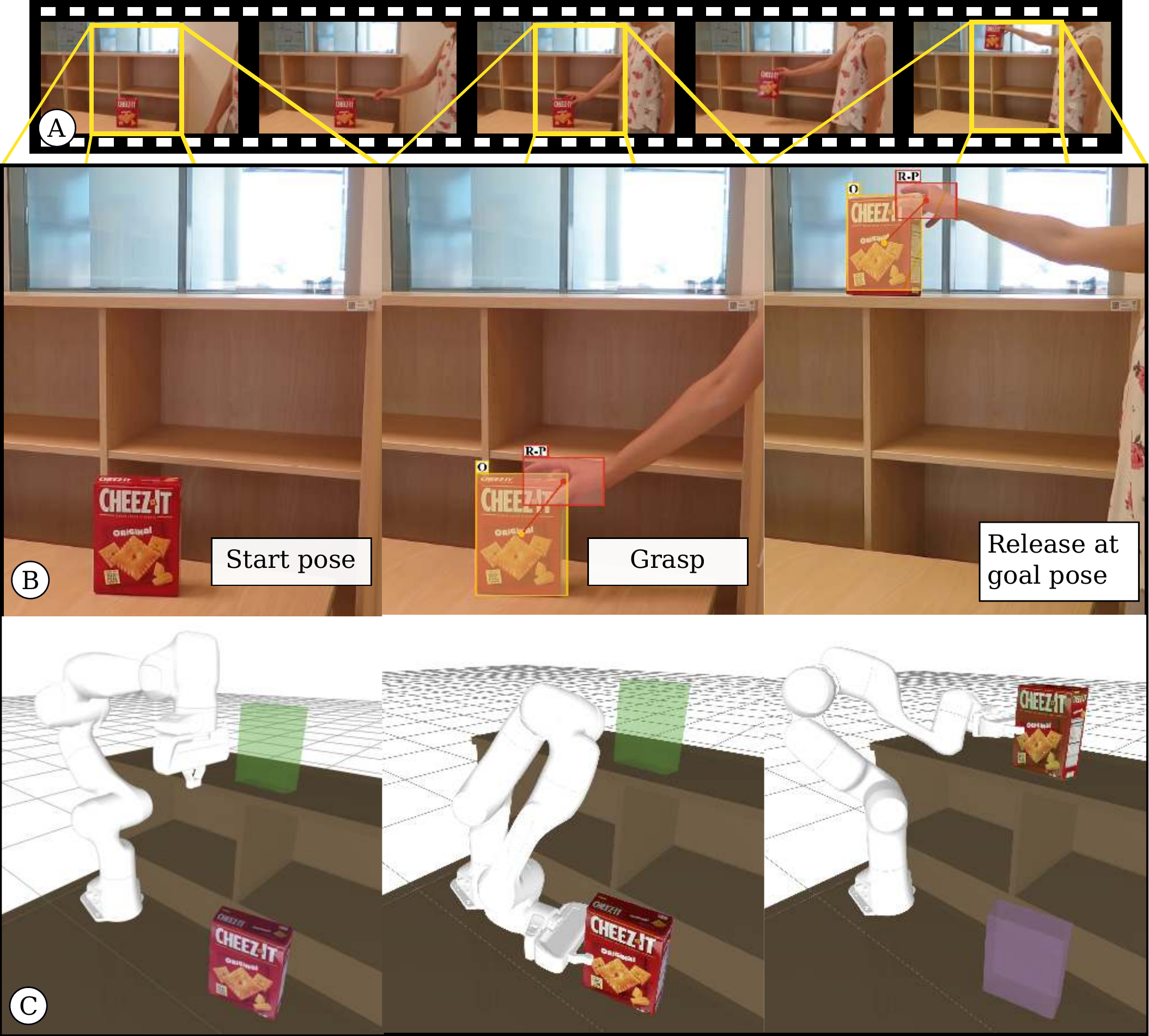}
  \caption[Multi-step task and motion planning.]{
  The proposed planning approach is guided by the demonstration video~(A). 
  The video depicts a person manipulating a known object; the \textit{cheez-it} box in this particular example.
  The video can contain several pick-and-place actions with multiple objects. Here, only a short clip with only one object and one action is shown.  
  From the video, we recognize (i)~the contact states between the human hand and the object, marked by red bounding boxes in~(B); and (ii)~the object 6D pose (3D translation and 3D rotation w.r.t camera) at the grasp and release contact states, marked in yellow in~(B). The robot trajectory planned by the proposed approach is shown in~(C). 
  The start and goal object poses in (C) are shown in magenta and green, respectively.
    }

  \label{fig:teaser}
\end{figure}

\section{Introduction}
Traditional robot motion planning algorithms seek a collision-free path from a given starting robot configuration to a given goal robot configuration~\cite{lynch2017modern}.
Despite the large dimensionality of the configuration space, sampling-based motion planning algorithms~\cite{KSLO95,rrt_connect} have shown to be highly effective for solving complex motion planning problems for robots, ranging from six degrees of freedom (DoFs) for industrial manipulators to tens of DoFs for humanoids~\cite{kuffner2005motion}.
Manipulation task-and-motion planning~(TAMP)~\cite{lamiraux2021hpp} adds an additional complexity to the problem by including movable objects in the state space. This requires the planner to discover the pick-and-place actions that connect the given start and goal robot configurations, bringing the manipulated objects from their start poses to their goal poses.
Planning long sequential tasks with continuous robot motion and discrete changes in contact (grasping and releasing the object) is recognized to be a challenging problem.

To address this challenge, we propose to leverage human demonstrations of the given task provided by an instructional video.
Changes of contacts, which are difficult to discover by planning, are extracted from the instructional video together with the 6D poses of \textit{a priori} known objects as visualized in Fig.~\ref{fig:teaser}.
The extracted information is then used to guide the proposed RRT-like algorithm that simultaneously grows multiple trees around the grasp and release states recognized in the video. The planned trajectory is finally locally optimized with an OCP solver to produce a smooth robot movement.
To show the benefits of guidance by an instructional video, we design a new benchmark with three challenging multi-step tasks for manipulation TAMP inspired by real-life problems and difficult to solve otherwise by state-of-the-art TAMP solvers.

\begin{figure*}[t]
  \centering

    \includegraphics[width=0.99\linewidth]{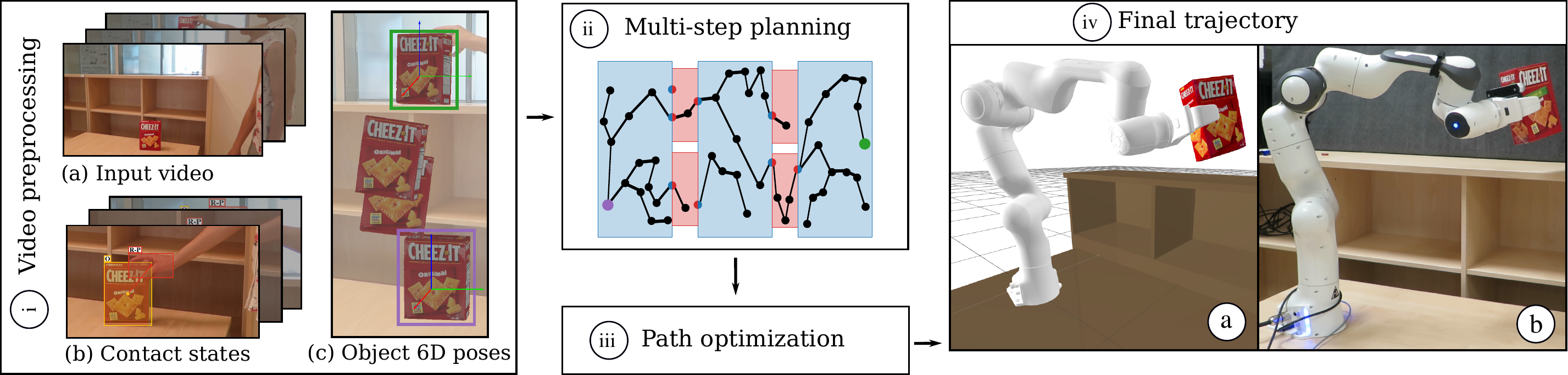}

  \caption[Overview of the ]{
  {\bf Approach overview.} (i)~First, we extract contact states and 6D object poses from the input instructional video, as described in Sec.~\ref{sec:extracting_contacts_poses}. (ii)~Next, we grow multiple trees in the admissible configuration space until we find a path between the start and goal configurations. More details on the state space are in Sec.~\ref{sec:state_space}, and more details on planning the path in Sec.~\ref{subsec:plan_between_states}. (iii)~This path is then further refined by an optimization module and (iv) executed either in simulation~(iv-a) or on a real-world robot~(iv-b).
  }

  \label{fig:pipeline}
\end{figure*}


The contributions of this paper are three-fold:
\begin{itemize}
    \item we propose an RRT-like multi-tree planner that incorporates video demonstration to solve complex multi-step TAMP tasks (summarized in Fig.~\ref{fig:pipeline});
    \item we design the trajectory refinement procedure that uses an OCP solver to find a smoother robot trajectory while avoiding collisions;
    \item we design a new benchmark with three challenging tasks to facilitate progress on this challenging problem;
    \item we provide experimental results demonstrating that our approach outperforms state-of-the-art TAMP solvers on the proposed benchmark.
\end{itemize}

This paper is an extended version of our previously published conference paper~\cite{my_icra_paper}. We extend it by investigating the generalization capabilities of the developed planning approach to new object poses, objects, and environments. We also introduce an additional trajectory refinement procedure formulated as an optimal control problem. This allows faster and smoother final trajectories for execution on the real robot.
We refer to the project page~\cite{project_page} for the supplementary video and code.

\section{Related work}

\noindent\textbf{Motion planning.} 
A popular choice for planning the motion of robotic manipulators is to use sampling-based planners inspired by RRT~\cite{lavalle1998rrt}, for example, RRT connect~\cite{rrt_connect} that grow trees simultaneously from the start and goal configurations.
Our approach is also inspired by RRT but simultaneously grows multiple trees.
Using multiple trees for motion planning has been explored in, for example, ~\cite{local_trees_rrt, multi_goal_rrt, multi_tree_cost_rrt}.
In~\cite{local_trees_rrt}, two categories of trees are used: (i)~global trees, that are initialized at the start and goal configurations and~(ii) local trees, that are initialized from configurations that fail to connect to global trees.
Local trees have a higher probability of being constructed in difficult regions such as narrow passages.
The trees grow independently until merged with each other. 
Authors of~\cite{multi_goal_rrt} and~\cite{multi_tree_cost_rrt} explore multi-tree approaches for solving multi-goal planning problems by constructing a tree for every target location.
%
%
In our approach, the additional trees are sampled at contact state changes that are hard to explore by the planner alone.
In our case, the input video demonstration provides the time of the contact state change and the 6D poses of objects for which we sample the 'local' trees. This provides a significant advantage compared to the other sampling methods. 

\noindent\textbf{Manipulation task-and-motion planning} seeks the robot motion that manipulates objects to achieve their given goal poses.
This manipulation is often referred to as object rearrangement planning and was studied by several TAMP solvers~\cite{lamiraux2021hpp, Symbolic_planning_Garrett_2020, rearrange1, rearrange2, mc_rearrange_table_yann} with some approaches limiting their applicability to 2D tabletop manipulation~\cite{rearrange1, rearrange2, mc_rearrange_table_yann} where objects are picked and placed only on a given flat surface of a table.
Compared to these methods, we solve 3D rearrangement tasks where objects are moved outside the flat surface of the table, for example, onto a shelf.
An existing approach not limited to 2D rearrangement planning~\cite{Symbolic_planning_Garrett_2020} combines symbolic planning with conditional samplers and is effective for multi-step and cost-sensitive problems.
However, it still suffers from the limitations of sampling-based methods. For example, discovering narrow passages in the configuration space remains a major open challenge. 
In the proposed work, the input instructional video guides the planner through these narrow passages in the configuration space.
Another general TAMP solver is called \textit{Humanoid Path Planner} (HPP)~\cite{lamiraux2021hpp} and uses RRT connect~\cite{rrt_connect} to solve TAMP tasks.
The planning in HPP is performed in admissible configuration space that is represented by a set of manifolds (denoted \textit{states} in HPP) defined by placement and/or grasp numerical constraints.
The movement between the states is achieved through \textit{transitions} that allow \eg~to grasp the object at a specific handle.
We build on the HPP planner and construct an admissible configuration space based on the input video demonstration.


\noindent\textbf{Guiding planners by demonstration} without using video has been explored in several set-ups.
For example, several kinesthetic demonstrations obtained by operators guiding the robot manually are used in~\cite{guided_motion_plan} together with a sampling-based planner to solve household tasks like transferring sugar with a spoon or cleaning a table. 
Demonstrations in 3D simulation obtained by manipulating a 22~DoFs glove are used in~\cite{grasp_from_demo_2011} to provide a sequence of actions and to identify parts of the objects that are grasped.
The planner is finding a collision-free path for the robot to repeat the sequence of pick-and-place actions. 
Instead of obtaining demonstrations by kinesthetic guidance or teleoperation by a glove, we propose to extract demonstrations from a video depicting the human performing the given task.  
Related to us, a video of a human demonstration has been used to obtain a sequence of commands for a robot in~\cite{nguyen2018translating}. It is assumed that a robot controller is available to perform these commands, for example, to grasp or to carry an object, or to pour from the cup.
Therefore only the task-space action planning is performed. 
In our approach, we perform task-and-motion planning and do not rely on given internal controllers to perform the individual manipulation actions.
Instead, we rely on a contact recognizer~\cite{contacts} and a 6D pose estimator~\cite{cosypose} to extract the relevant guidance for the planner from the video.
Such contact and object pose estimators from images have been shown to be robust to various lighting conditions, camera properties, etc.
While there are several methods to estimate finger-level grasps of objects~\cite{hand_grasp_fingers, hand_grasp_fingers2}, robots do not typically grasp objects in the same way as humans and therefore the robotic planner may not benefit from this information. Hence, we focus only on information related to the pose of objects and the contact states and bypass completely the challenging task of extracting finger-level grasping information from the input video. 

\noindent\textbf{Learning-based methods.} 
Policy search methods, for example, reinforcement learning, are widely used to solve manipulation tasks~\cite{levine2016end, deisenroth2011learning, ennen2019learning}.
However, these methods usually require substantial training time, and once the policy is learned, it usually does not generalize well to different scenes, for example, if the furniture is moved. 
To avoid these limitations, we rely on a geometry-based planning approach that is able to re-plan the path if the geometry of the scene changes.

\begin{figure*}[t]
  \centering

    \includegraphics[width=\linewidth]{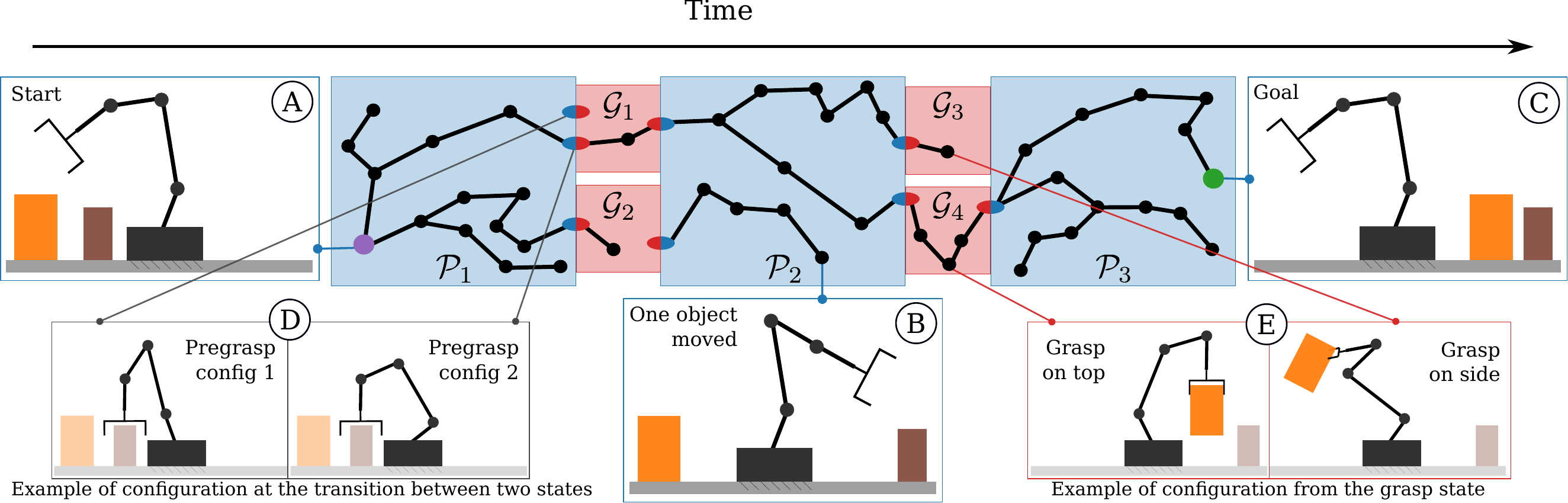}

  \caption[Configuration space]{
 \bf{The admissible configuration space} is a set of (i) \textcolor{blue_tab}{placement states}~$\mathcal{P}_i$ where objects rest on the contact surfaces at specific poses and (ii) \textcolor{red_tab}{grasp states}~$\mathcal{G}_j$ where one of the objects is grasped by the robot gripper. To transit to another \textcolor{blue_tab}{placement state}, \ie~to change the pose of one object, the robot needs to grasp the object and release it at the new location. Since we define several ways to grasp each object, we can transition through multiple \textcolor{red_tab}{grasp states} to build a path between configurations given by two consecutive \textcolor{blue_tab}{placement states}.
   Start configuration~(A) lies in the first placement state $\mathcal{P}_1$.
   To achieve the goal, the robot needs to move the brown object first, \ie~to reach state~$\mathcal{P}_2$~(\eg~configuration~(B)).
   Finally, the robot moves the second object (reach~$\mathcal{P}_3$) and moves robot configuration to the given goal~(C).
  The configuration at the transition between the~\textcolor{blue_tab}{placement state} and~\textcolor{red_tab}{grasp state} is not unique as there are various robot joint values resulting in the same 'pregrasp' poses as shown in~(D).
Multiple ways of grasping the same object are represented by different \textcolor{red_tab}{grasp states} as shown in~(E). The video demonstration provides object poses~(see Sec.~\ref{sec:extracting_contacts_poses}) to construct the placement states $\mathcal{P}_1$, $\mathcal{P}_2$, $\mathcal{P}_3$~(see Sec.~\ref{sec:state_space}). Finally, the planner spawns and expands multiple trees~(see Sec.~\ref{subsec:plan_between_states}) until it finds the path between start and goal configurations.
}

  \label{fig:manifolds}
\end{figure*}

\section{Multi-step planning guided by video demonstration}

The proposed method uses a video demonstration to guide the TAMP algorithm for the given manipulation task.
The overall pipeline is visualized in Fig.~\ref{fig:pipeline}. 
The input to the proposed approach is the instructional video depicting a human manipulating objects and the geometrical description of the scene.
The output of the approach is a collision-free path for the robot that solves demonstrated manipulation task.

\subsection{Extracting contact states and object poses from input video}

\label{sec:extracting_contacts_poses}
The objective of the first stage is to extract the changes of the contact states, \ie~when a human grasps or releases an object, and extract the object 6D poses at the times of the contact state changes.
The input video demonstrates a human manipulating known objects one by one from their starting poses toward their goal poses.
We focus the proposed method on a single-arm robotic manipulator; therefore, the human demonstrator only moves one object at a time.
We apply the hand contact recognizer~\cite{contacts} to obtain a sequence of contact states $c_k \in \{ \text{grasp}, \, \text{release}, \, \text{none} \}^N$, where $N$ is the number of movable objects, the contact changes are indexed by $k \in \{1, \ldots, T \}$ with $T$ being the number of changes in contact states. 
Therefore, the variable~$c_k$ contains the contact state for each object in the scene at the time of contact state change~$k$.
The contact states of the non-manipulated objects are set to 'none'.
The 3D models of objects used in the demonstration are known, which is a reasonable assumption in many practical robotics setups, such as in manufacturing. 
As a result, the object 6D poses can be estimated from a single video frame by an off the shelf pose estimator such as CosyPose~\cite{cosypose}.
The 6D pose is estimated for every movable object in the scene at the time of the contact state change~$k$.
The poses of the objects are denoted as \( A_{k,i} \in SE(3) \),  where \( k \) represents the contact change index and \( i \) represents the object index. 
Let \( \mathcal{A}_k \) represent the set of poses of all objects at the contact change \( k \), \ie, \( \mathcal{A}_k = \{ A_{k,1}, A_{k,2}, \ldots, A_{k,N} \} \).
We use the recognized contact states and object poses to constrain the configuration space in which the planning is performed. 

\subsection{The configuration space built from contacts and poses}
\label{sec:state_space}
We now explain the topology of the space where the planner searches for the robot trajectory and how it connects to the demonstration.
In the case of manipulation TAMP, planning is performed in an admissible configuration space, which is defined by a set of numerical constraints that prevent objects from flying in the air or moving without being grasped by the robot.
Following~\cite{lamiraux2021hpp}, we represent the space of admissible configurations by a set of states, \ie~a set of manifolds, in which the admissible configurations lie.
Each state is defined by numerical constraints that must be satisfied for the configuration that belongs to it.
Here, a configuration is defined by the position and orientation of objects in the scene as well as the configuration of the robot.  
The configurations that belong to each state are subject to placement and/or grasp constraints. 
The placement constraint for an object enforces that the object remains in the same stable pose on the contact surface. 
A grasp constraint for an object enforces that the object remains static \wrt~the gripper frame. 
For each object, we define several possible handles, \ie~several transformations between the robot gripper and the object. 
We define \textit{placement states} as those that are subject to only placement constraints and \textit{grasp states} as those that are subject to the grasp constraint for one of the objects and placement constraints for the rest of the objects. 
The transition between the placement and the grasp states represents configurations in which an object is grasped but lies on the contact surface at the same time, \ie~constraints from the placement and grasp states are satisfied simultaneously.
We call such states \textit{neighboring states}.

The challenge of TAMP algorithms lies in the dimensionality of the search space, for example, the planning algorithm has to explore many possibilities for placing the object on the contact surface. 

Here, we propose to guide the exploration by selecting only the placement states observed in the demonstration video. 

For each contact change $k$, we use poses of all objects~$\mathcal{A}_k$ to construct the placement state $\mathcal{P}_k$ (blue in Fig.~\ref{fig:manifolds}).
The constructed placement states represent the sequence of object placements observed in the video.
The consecutive placement states are \textit{connected} by grasp states (red in Fig.~\ref{fig:manifolds}) as the object needs to be grasped and moved to change the placement state, \ie~to change the pose of one object.
There are multiple handles for each object, and therefore there are multiple grasp states, \ie~multiple ways to grasp a specific object.
An example of such an admissible configuration space is shown in Fig.~\ref{fig:manifolds}, where three placement states are created:
the first placement state~$\mathcal{P}_1$ is constrained (given) by the initial poses of objects,
the next placement state~$\mathcal{P}_2$ is constrained by the starting pose of the orange object and the goal pose of the brown object,
and the last placement state~$\mathcal{P}_3$ is constrained by the goal poses of both objects.
In the example, two handles were defined for each object, representing (i)~grasp on top and (ii)~grasp on the side.
For the path to be feasible, it must contain only the configurations from the admissible space and satisfy additional constraints of the environment, \eg~robot configuration must be within the joint limits, and the configuration must be collision-free.
We refer to the space in which the configurations satisfy all restrictions as $\mathcal{C}_\text{free}$.

\begin{algorithm}
\caption{multi-step RRT guided by demonstration}\label{alg:multi_contact_rrt}
\begin{algorithmic}[1]
    \Require Contact states $c_k$,
    object 6D poses $\mathcal{A}_k$, 
     $\bm q_\text{start}$, $\bm q_\text{goal}$, $\eta_\text{sample\_tree}$, step size $\delta$

    \State $\mathcal{T} \gets \{ \text{init\_tree}(\bm q_\text{start}), \text{init\_tree}(\bm q_\text{goal})\}$ \Comment{Existing trees}
    
    \Repeat
        \State $p \sim \mathcal{U}(0,\, 1)$
        \If{$p < \eta_\text{sample\_tree}$} \Comment{Sampling a new tree}
            \State $k \sim \{1,\ldots,T \}$
            \State $\bm q_\text{from}, \bm q_\text{to} \gets \text{sample\_on\_transition}(c_k,\, \mathcal{A}_k)$
            \If{$\bm q_\text{from} \notin C_\text{free}$ \textbf{or} $\bm q_\text{to} \notin C_\text{free}$}
                \State \textbf{continue}
            \EndIf
            \State $t \gets \text{init\_tree}(\bm q_\text{from})$
            \State $t.\text{add\_edge}(\bm q_\text{from},\, \bm q_\text{to})$
            \State $\mathcal{T} \gets \mathcal{T} \cup \{t\}$
            \State $\text{attempt\_link}(\bm q_\text{from},\, \mathcal{T}, \, \delta)$ \Comment{Alg.~\ref{alg:attempt_link}}
            \State $\text{attempt\_link}(\bm q_\text{to},\, \mathcal{T}, \, \delta)$ \Comment{Alg.~\ref{alg:attempt_link}}
        \Else \Comment{Tree growing}
            \State $\mathcal{S} \gets \text{sample\_state}(\mathcal{T})$
            \State $\bm q_\text{rand} \gets \text{sample\_configuration}(\mathcal{S})$
            \State $\bm q_\text{nn} \gets \text{nearest\_neighbor}(\bm q_\text{rand},  \,  \mathcal{S},  \,  \mathcal{T})$
            \State $\bm q_\text{step} \gets \bm q_\text{nn} + \delta(\bm q_\text{rand} - \bm q_\text{nn})$
            \If{$\bm q_\text{step} \notin C_\text{free}$}
                \State \textbf{continue}
            \EndIf
            \State $t \gets \text{get\_tree}(\bm q_\text{nn})$
            \State $t.\text{add\_edge}(\bm q_\text{nn},  \,  \bm q_\text{step})$
            \State $\text{attempt\_link}(\bm q_\text{step}, \, \mathcal{T},\, \delta )$ \Comment{Alg.~\ref{alg:attempt_link}}
        \EndIf
    \Until $\text{get\_tree}(\bm q_\text{start}) = \text{get\_tree}(\bm q_\text{goal})$ or out of resources
\end{algorithmic}
\end{algorithm}

\subsection{Planning between contact states}
\label{subsec:plan_between_states}
We now explain how the RRT algorithm explores the admissible configuration space defined in the previous section.
Transitioning between the placement and grasp states, which is necessary to move the objects, remains a challenge for sampling-based planners such as RRT~\cite{rrt_connect} as it needs to sample configurations that satisfy constraints from both neighboring states.
To address that issue, we design an extension of RRT that grows multiple trees simultaneously with tree roots sampled at the transitions between placement and grasp states.
The overview of the proposed algorithm is shown in~Alg.~\ref{alg:multi_contact_rrt}.
The algorithm is split into two main routines: (i)~sampling of a new tree at the transition between the placement and grasp states, and (ii)~growing an existing tree at a randomly selected state.
We randomly select which routines are used in each iteration with a Bernoulli distribution controlled by a parameter $\eta_\text{sample\_tree}$.
The algorithm stops if both start and goal configurations are connected into a single tree or if we run out of resources (\eg, the maximum number of iterations or maximum planning time).

\noindent\textbf{Sample from the state.} One of the main capabilities required by the algorithm is sampling from the given state.
Similarly to~\cite{lamiraux2021hpp}, we sample a random configuration from Euclidean space and call a numerical solver~\cite{NoceWrig06} to compute iteratively the configuration that satisfies the numerical constraints of the state.
To sample the transition connecting two states, the constraints from both states are concatenated.
However, to assign a unique state to each configuration sampled from the transition, we construct two identical configurations denoted~$\bm q_\text{from}$ and $\bm q_\text{to}$ and assign states to them.
The two main routines of Alg.~\ref{alg:multi_contact_rrt}, which use the described sampling procedure, are discussed next.

\noindent\textbf{Sampling a new tree at the transition} between the placement and grasp states is performed as follows. 
First, we sample a pair of configurations ($\bm q_\text{from}, \bm q_\text{to}$) that satisfy the constraints of the randomly selected transition.
If both configurations also satisfy the constraints of the environment (\ie~respect the joint limits and are collision-free), we create a new tree containing a root ($\bm q_\text{from}$) and a leaf ($\bm q_\text{to}$).
We attempt to link both the created configurations to the existing trees in their corresponding states.

\noindent\textbf{Tree growing} in a randomly selected state is performed as follows. This routine consists of sampling a new configuration in admissible configuration space and extending the tree in the direction of the sampled configuration.
We randomly choose a state~$\mathcal{S}$ and sample the random configuration~$\bm q_\text{rand}$ from it.
The nearest neighbor of the sampled configuration~$\bm q_\text{nn}$ is found such that~$\bm q_\text{nn}$ also lies in the state~$\mathcal{S}$.
A new configuration~$\bm q_\text{step}$ is computed along the segment between $\bm q_\text{nn}$ and~$\bm q_\text{rand}$ in the manually defined step-size distance $\delta$ from~$\bm q_\text{nn}$.
If $\bm q_\text{step}$ is collision-free, we add it to the tree.
Finally, we attempt to link the new configuration to existing trees with nodes in the same state.

\begin{algorithm}
\caption{{\bf Attempt to link} function connects a given configuration to other trees}\label{alg:attempt_link}
\begin{algorithmic}[1]
\Require configuration $\bm q$, set of trees $\mathcal{T}$, step size $\delta$,
    \State $\mathcal{S} \gets \text{get\_state}(\bm q) $
    \For{$t \in \mathcal{T} \setminus \text{get\_tree}(\bm q) $}
        \If{t has nodes in $\mathcal{S}$}
            \State $\bm q_\text{nn} \gets \text{nearest\_neighbor}(\bm q, \, \mathcal{S}, \, \{t\})$
            \State $\bm q_\text{parent} \gets \bm q_\text{nn}$
            \For{$\bm q_\text{step} \in \{\bm q_\text{nn} + \delta(\bm q - \bm q_\text{nn}), \ldots ,  \bm q\}$}
                \If{$\bm q_\text{step} \notin C_\text{free}$}
                    \State \textbf{break}
                \EndIf
                \State $t.\text{add\_edge}(\bm q_\text{step}, \, \bm q_\text{parent})$
                \State $\bm q_\text{parent} \gets \bm q_\text{step}$
            \EndFor
            \If{$\bm q_\text{step} = \bm q$}
                \State merge $t$ and tree containing $\bm q$
            \EndIf
        \EndIf
    \EndFor
\end{algorithmic}
\end{algorithm}

\noindent\textbf{Attempt to link} function is used by Alg.~\ref{alg:multi_contact_rrt} to connect given configuration~$\bm q$ with the existing trees. The function is summarized in Alg.~\ref{alg:attempt_link}.
It searches for a linear path between the configuration~$\bm q$ and other tree nodes in the same state.
For each tree, the nearest neighbor~$\bm q_\text{nn}$ is found.
Then, configurations along the segment from~$\bm q_\text{nn}$ to $\bm q$ are added to the tree unless a collision is observed. 
If the entire path is collision-free, we merge the trees that contain the configurations $\bm q_\text{nn}$ and $\bm q$.

\begin{figure*}[!t]
\centering
\hfil
\subfloat[]{\includegraphics[width=0.3\textwidth]{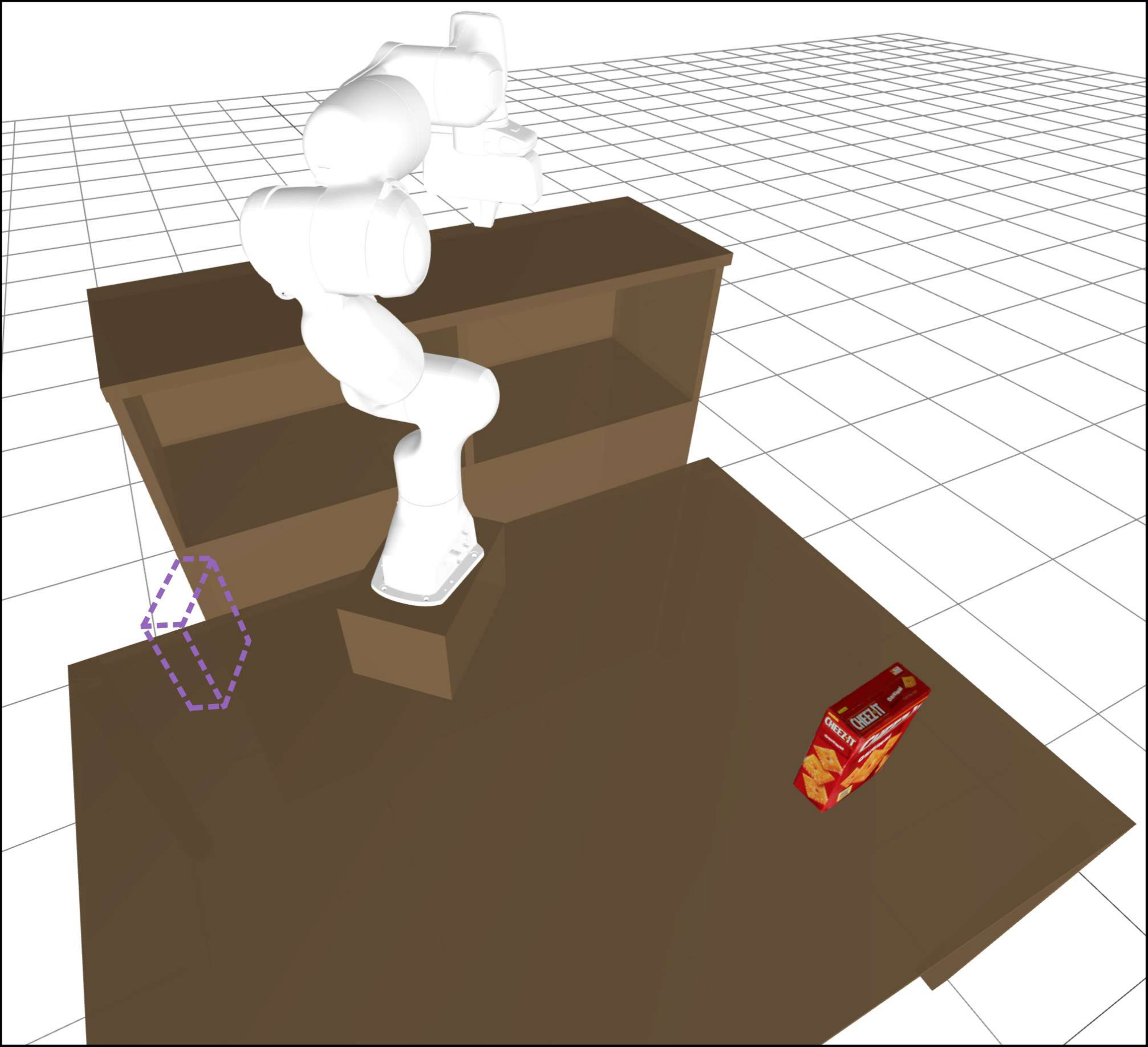}%
\label{subfig:generalization_object_start}}
\hfil
\subfloat[]{\includegraphics[width=0.3\textwidth]{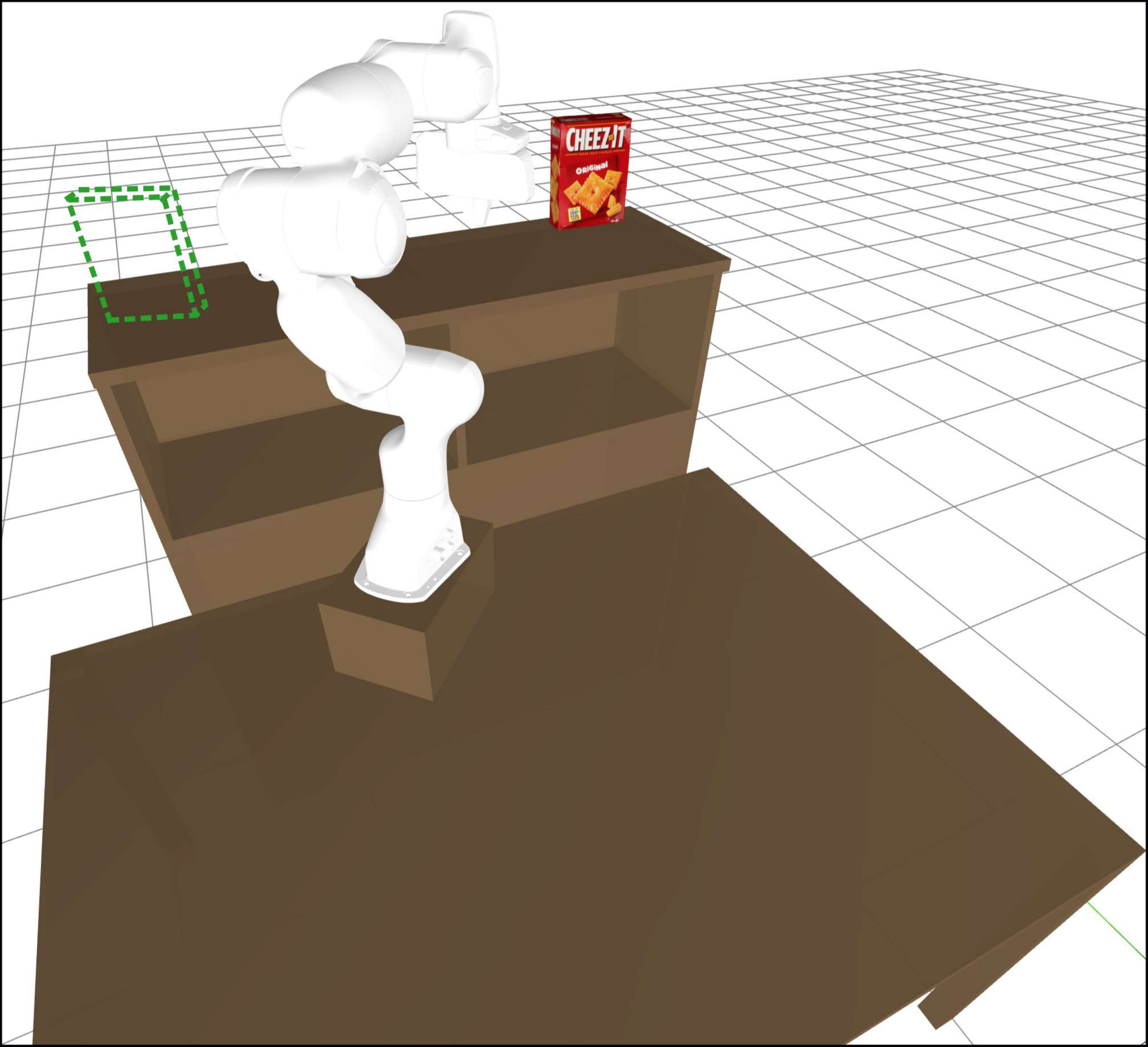}%
\label{subfig:generalization_object_goal}}
\hfil
\subfloat[]{\includegraphics[width=0.284\textwidth]{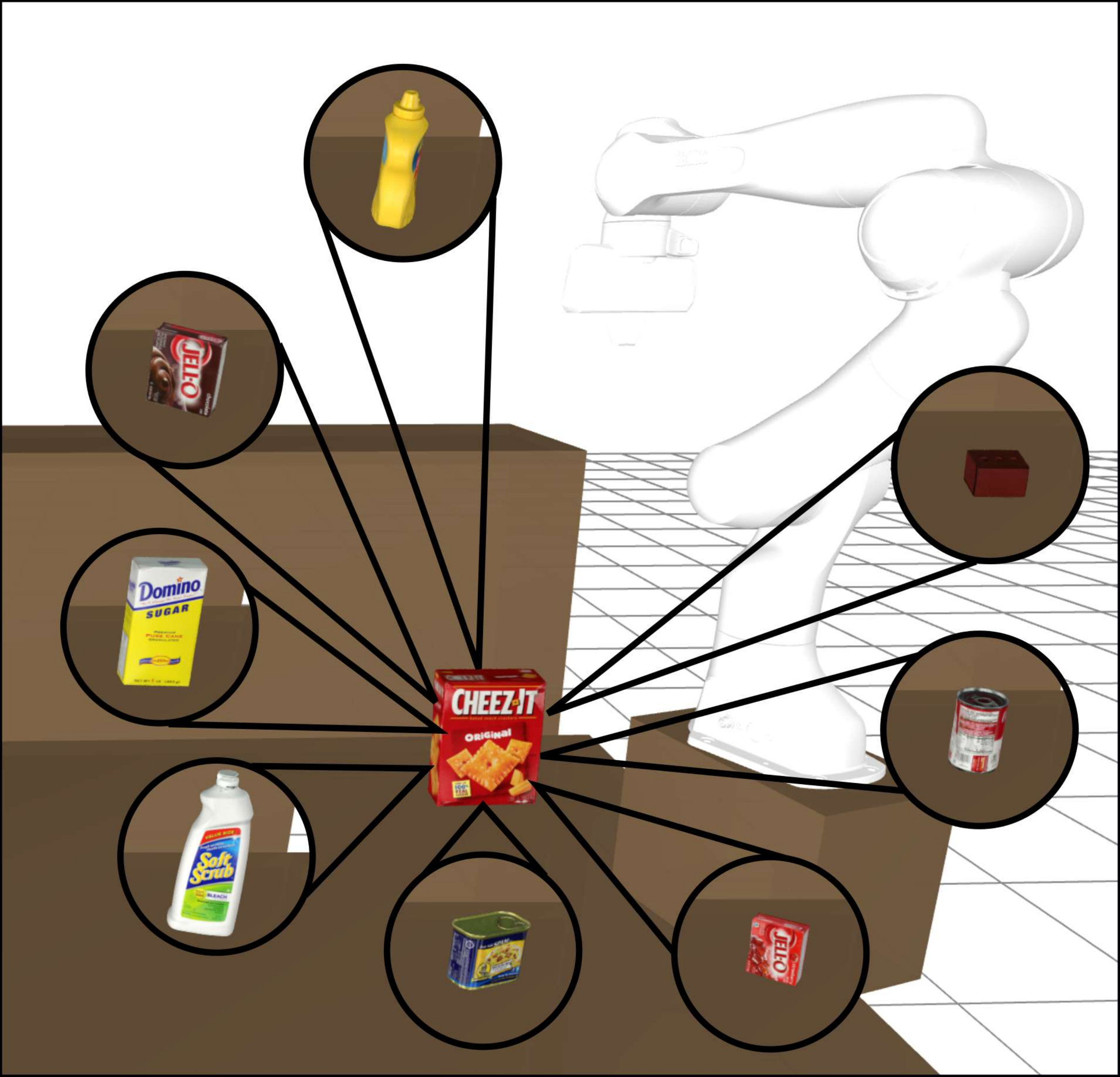}%
\label{subfig:all_possible_objects}}
\hfil
\\
\centering
\subfloat[]{\includegraphics[width=0.9\textwidth]{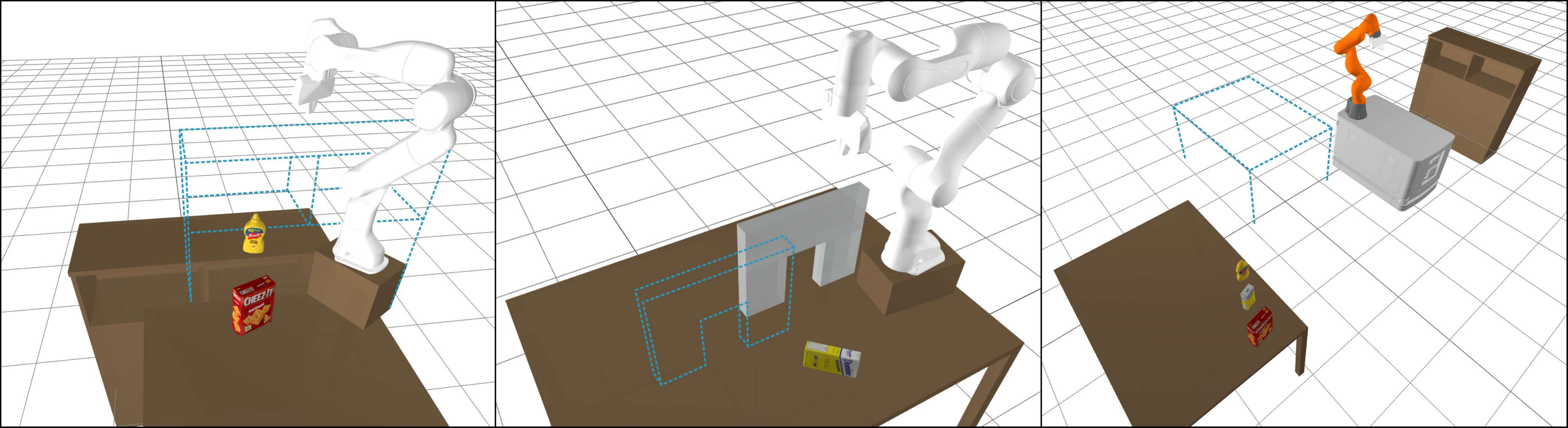}%
\label{subfig:generalization_env}}

  \caption[Generalization scenarios.]{
  {\bf Generalization scenarios} include variation of: start object poses (\ref{subfig:generalization_object_start}), goal object poses (\ref{subfig:generalization_object_goal}), object type (\ref{subfig:all_possible_objects}) and environment (\ref{subfig:generalization_env}).
  {\bf \ref{subfig:generalization_object_start}, \ref{subfig:generalization_object_goal}: start and goal object pose variation.} We sample randomly $x$ and $y$ displacements for object start (goal) poses and a rotation around $z$ axis. {\bf \ref{subfig:all_possible_objects}: Object variation.} We illustrate the variability of objects that our method supports. {\bf \ref{subfig:generalization_env}: Environment variation.} Furniture pose variability for {\bf shelf}, {\bf tunnel}, and {\bf waiter} tasks. We sample randomly $x$ and $y$ displacements for the tunnel in the {\bf tunnel} task; $x$, $y$, and $z$ axis displacements for the shelf in the {\bf shelf} task and the table in the {\bf waiter} task. }
    
\label{fig:generalization}
\end{figure*}

\subsection{Generalization to new poses, objects, and environments}
\label{subsec:generalization}
Robotic systems must operate in dynamic and unpredictable environments where object poses, object types, and environmental layouts can vary significantly. 
Generalizing from a single demonstration to various scenarios is crucial for creating flexible and robust algorithms that can adapt to new tasks without requiring a new demonstration.
To go beyond the demonstrated scenario, we investigate the generalization capabilities of our approach across diverse setups.

\noindent \textbf{Start and goal object pose variation.}
After extracting the object poses $\mathcal{A}_k$ from the demonstration~(Sec.~\ref{sec:extracting_contacts_poses}), we can replace the start object poses~$\mathcal{A}_0$ or the goal object poses~$\mathcal{A}_T$ to plan for a new scenario. 
This approach allows us to update the demonstration to use it for novel start and goal poses during execution.
For example, the initial poses of objects can be detected by the CosyPose pose estimation method~\cite{cosypose}.

\noindent \textbf{Object variation.} 
The post-processed demonstration contains the poses of the objects that can also be used for different geometries as the planner adjusts to the geometrical properties of the object.
Therefore, a single demonstration can be used to manipulate various geometries of objects.

\noindent \textbf{Environment variation.} 
In our formulation (Sec.~\ref{sec:extracting_contacts_poses}), all object poses are expressed in a common world frame of reference.
To accommodate for changes in the pose of the furniture, we associate each object pose~$A_{k,i}$ with a specific frame of reference (typically attached to a contact surface, \eg, the table).
This allows us to modify the poses of objects in demonstration based on the poses of the furniture and therefore to accommodate for environment variation.

\subsection{Path optimization sing random shortcut} 
\label{sec:path_optimization}
The proposed multi-step RRT produces a feasible path in the admissible configuration space that achieves the goal.
However, this path often contains redundant motion, for example, approaching an object several times before grasping it.
To further optimize the path, we apply an approach similar to~\cite{random_shortcut_example}: we select two random configurations $\bm q_1$ and $\bm q_2$ that lie on the path and belong to the same state. We try to build a collision-free segment between them. If such a segment exists, we replace the portion of the path between $\bm q_1$ and $\bm q_2$ by this segment.

\subsection{Refine trajectory by solving an optimal control problem}
\label{subsec:trajopt}
While the planned trajectory provides a feasible path, it often results in joint velocities that are not smooth, which is undesirable for robot control. To address this, we introduce an additional refinement of the trajectory. This is a challenging problem, as it requires optimizing for accurate yet smooth motions while also avoiding collisions. To address this challenge, we formulate an optimal control problem to find control torques for the robot~\cite{aligatorweb, jalletPROXDDPProximalConstrained2023}.

In detail, to refine the trajectory, we define keyframes as the time indices in which the configuration lies in the intersection of manifolds (\ie~pregrasp configuration, where the robot gripper is approaching the object).
The set of corresponding discrete-time indices is denoted by $\mathcal{K}$.
The end effector poses are computed on each keyframe and are denoted $F_{i}$, where $i \in \mathcal{K}$.
Our goal is to optimize the robot trajectory between consecutive keyframes so that the keyframe pose is reached with minimal motion while avoiding collisions.

We formulate the following trajectory optimization problem to find the sequence of length $H$ of control torques $\bm{u_0}^*,\ldots,\bm{u_H}^*$ that follow the grasp sequence of the planned trajectory:
\begin{equation}
\label{eq:optimization_problem}
\begin{aligned}
\min_{\bm u_0, \dots, \bm u_H} \displaystyle \sum_{i \in \mathcal{K}}{w_d d(\bm q_{i}, \bm F_{i}}) + {\sum_{t=0}^{H} L_t(\bm x_t, \bm u_t)} \\
    \text{s.t. } \bm x_t  = f(\bm x_{t-1}, \bm u_{t-1}) \quad \forall t = 1, \ldots, H
\end{aligned}
\end{equation}

where $\bm x_t = (\bm q_t, \bm v_t)$ is a state variable that concatenates robot joint configuration $\bm q_t$ and robot joint velocity $\bm v_t$, $f(\cdot)$ represents the dynamics of the robot, implemented with the Articulated Body Algorithm \cite{carpentier2019pinocchio}. Gravity is set to zero in $f(\cdot)$ since our experimental platform software uses internal gravity compensation. The initial state $\bm x_0$ is given by the planner.

\begin{figure*}[t]
    \centering
    \subfloat[]{\includegraphics[height=1.75in]{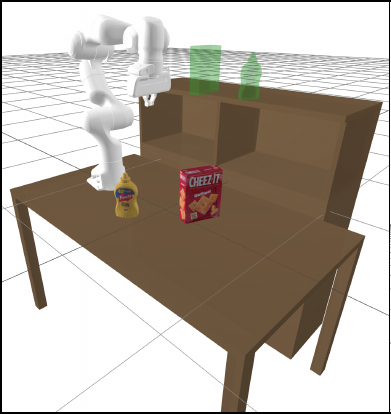}
    \label{benchmark_shelf}}
    \hfil
    \subfloat[]{\includegraphics[height=1.75in]{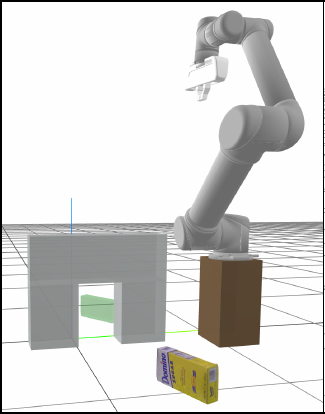}
    \label{benchmark_tunnel}}
    \hfil
    \subfloat[]{\includegraphics[height=1.75in]{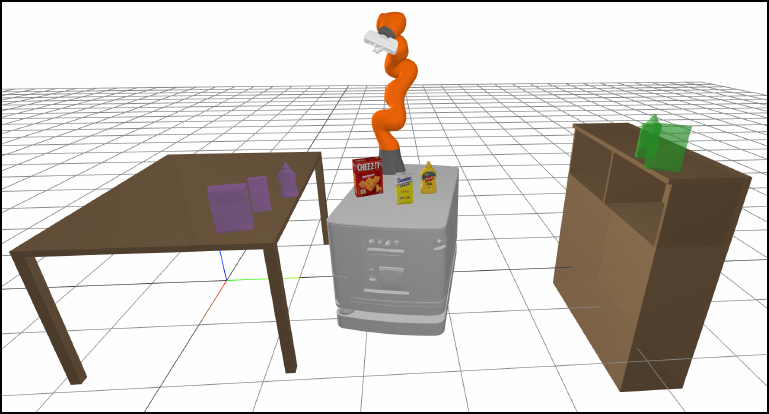}
    \label{benchmark_waiter}}
    \caption[Benchmark]
    {
  {\bf The proposed benchmark} includes three tasks: the shelf task~(a) with the goal of moving several objects to a predefined pose on the table or on the shelf; the tunnel task~(b) with the goal of transferring an object through the tunnel; the waiter task~(c) with the goal of moving several objects to a distant location while using the tray to minimize the traveled distance.

    }
    
  \label{fig:benchmark}
\end{figure*}

The OCP is defined by the cost that consists of two components. The sum over $i \in \mathcal{K}$ penalizes the end-effector pose at state transition, where $d(\bm q, \bm F$) denotes the squared norm of SE(3) $\log$~\cite{microlie} between the end-effector pose computed by forward kinematics at~$\bm q$ and the target end-effector pose $\bm F$ obtained from the planned trajectory (scaled by weight~$w_d$). The sum over $t=0 \ldots H$ features the running cost $L_t$ defined by:

\begin{equation}\label{eq:running_cost}
\begin{aligned}
L_t(\bm x_t, \bm u_t) =
    w_x  (\bm x_t -  \hat{\bm x}_t)^\top (\bm x_t - \hat{\bm x}_t) 
    + w_u \bm u_t^\top \bm u_t  & \\
    + w_c c(\bm q_t)
    + w_b b(\bm q_t) &
\end{aligned}
\end{equation}

where the term $(\bm x_t -  \hat{\bm x}_t)^\top (\bm x_t - \hat{\bm x}_t)$ regularizes the state (scaled by weight~$w_x$) where the value $\hat{\bm x}_t$ is obtained from the path returned by the planner,
the term $\bm u_t^\top \bm u_t$ regularizes the control torques (scaled by weight~$w_u$), 
function $c(\bm q)$ represents the collision cost at configuration~$\bm q$ (scaled by weight~$w_c$), 
function $b(\bm q)$ represents the joint limit cost at configuration~$\bm q$ (scaled by weight~$w_b$). 
We initialize the solver to follow the initial planned path (\ie, large weight $w_x$) and relax this constraint in the course of optimization.
The collision cost is computed as a barrier function of the penetration distance between the geometries present in the scene. Its gradient is computed by finite differences.
The joint limit cost is computed as a weighted sum of squared violations of the joint limits.

\section{Experiments}

In this section, we first describe the new benchmark for multi-step TAMP, then give the details of the evaluation of our approach against the baseline methods, then discuss the generalization capabilities and provide both quantitative and qualitative evaluation in multiple generalization scenarios. 

\subsection{Benchmark for multi-step task \& motion planning}
We designed a new benchmark consisting of three challenging tasks that involve multiple contact changes, \ie~object pick-and-place actions.
The objects used for manipulation were selected from  the YCBV dataset~\cite{xiang2018posecnn_ycbv} for which the 6D pose estimator~\cite{cosypose} is available.
The camera used for the video demonstration recording was calibrated both intrinsically and extrinsically with respect to the environment.
This allows us to express object poses and furniture poses in a common frame of reference.
Multiple robots with varying number of degrees of freedom are tested in each task, including the Panda robot and the Kuka IIWA robot mounted on a mobile platform.
For the purpose of benchmarking, we sample the attachment pose of the non-mobile robots to the desk randomly, such that all pick-and-place poses in the demonstration are reachable.
For real scenarios, the robot-to-desk attachment is given by the environment.
Next, we provide the description of the tasks. Please see the supplementary video for their visualization.

\noindent\textbf{The shelf task} is composed of a table, a shelf and a varying set of objects that the robotic manipulator is supposed to arrange, \ie~to move the objects to the predefined poses on the table or on the shelf (see Fig.~\ref{benchmark_shelf}).
The complexity of the task is controlled by the number of objects that the robot should arrange.
This task is challenging for state-of-the-art planners because it requires moving multiple objects in a single planning task.

\noindent\textbf{The tunnel task} consists of a tunnel and a single object that should be transferred through the tunnel (see Fig.~\ref{benchmark_tunnel}).
The tunnel is thin enough so that the robot can place an object inside the tunnel on one side and pick it up from the other side.
The challenge lies in the narrow passage in the admissible configuration space that needs to be discovered by the planner.

\noindent\textbf{The waiter task} simulates the job of waiter, in which a set of objects needs to be transferred from one location to another distant location (see Fig.~\ref{benchmark_waiter}).
Waiters use a tray to minimize the walked distance.
In our simulation, a mobile robot is equipped with a tray-like space that it can use for transferring objects.
Discovering the tray in the planning state-space is non-trivial, which makes this task challenging for the planners that do not utilize demonstrations.

\subsection{Evaluation}
\label{sec:experimetns_evaluation}
\noindent\textbf{Baselines.} We compare the proposed approach with the following baselines: (i)~RRT-connect implemented in HPP~\cite{lamiraux2021hpp} without and (ii)~with the path optimization using random shortcut, (iii)~PDDLStreams method~\cite{Symbolic_planning_Garrett_2020}, and (iv)~the proposed approach without and (v)~with the path optimization using random shortcut.
For PDDLStreams we set the goal requirement to match the final object poses for every object. 
The object grasp definitions are set to be consistent with the proposed approach. 


\noindent\textbf{Metrics.} To compare the proposed approach with the baselines, we solve the planning problem on the same machine and use the following metrics: (i)~the time to find a solution measured in seconds, (ii)~the number of object grasps and releases, \ie, the number of contacts, and (iii)~the length of the planned path.
The length of the planned path is evaluated as the sum of Euclidean distances computed between the consecutive robot configurations of the planned path.

\begin{figure*}[t]
  \centering

    \includegraphics[width=\linewidth]{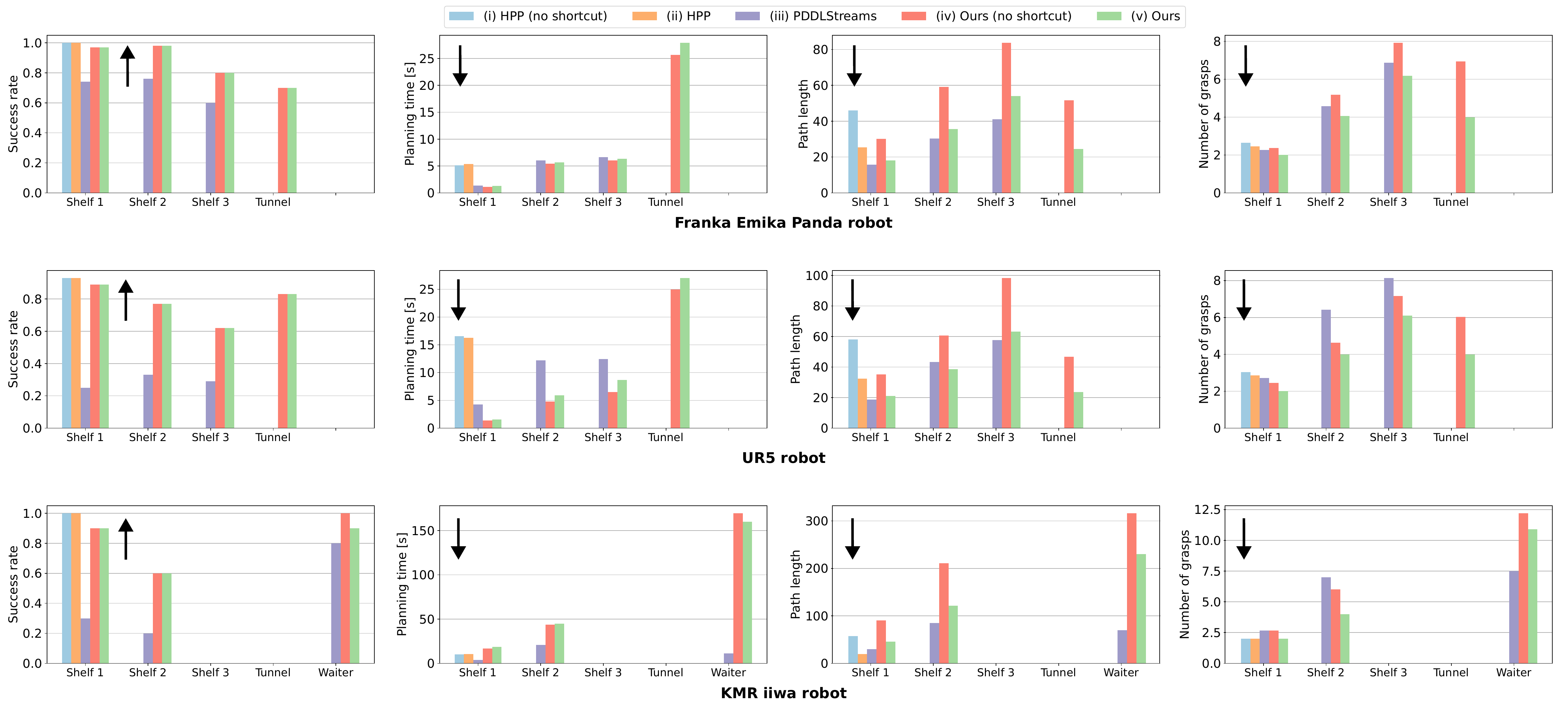}
  
  \caption[Results]{
  {\bf Results} reported for different robots (rows) and different metrics (columns).  We report (from left): the success rate, the planning time [s], the path length, and the number of grasps.
  For the success rate, the higher number the better ($\uparrow$); for the other metrics lower numbers are better ($\downarrow$).
  The plots show the comparison of our method with the following baselines: (i)~RRT-connect implemented in \textcolor{hpp_tab}{HPP}~\cite{lamiraux2021hpp} without and (ii)~ \textcolor{hpp_opt_tab}{with the path optimization}, (iii)~\textcolor{pddl_tab}{PDDLStreams} planner~\cite{Symbolic_planning_Garrett_2020}, and (iv)~our \textcolor{our_tab}{proposed approach} without and (v)~\textcolor{our_opt_tab}{with the proposed path optimization}. Each graph shows results for five different tasks. ``Shelf 1" - ``Shelf 3" correspond to the Shelf tasks with 1-3 objects. ``Tunnel" corresponds to the tunnel task and ``Waiter" corresponds to the waiter task. Please note that the Waiter task is only performed with the KMR iiwa mobile robot.  
    }

  \label{fig:restults}
\end{figure*}
\noindent\textbf{Results} are shown in Fig.~\ref{fig:restults}.
We set the time limit to 60~s for all tasks except the waiter task where the limit was set to 300~s.
Planning was repeated 10~times for KMR robot, and 10~times for each of~10 robot base poses for arm robots. The success rate is reported in the first column of Fig.~\ref{fig:restults}. 

For the successful runs, we measure the aforementioned metrics and report the average.
The different columns in Fig.~\ref{fig:restults} show the different reported metrics.
The different rows in Fig.~\ref{fig:restults} correspond to experiments with different robots (from top): (i) 7~DoFs Franka Emika Panda robot, (ii) 6~DoFs UR5 robot, and (iii) 7~DoFs KUKA IIWA arm mounted on 3~DoFs mobile platform~(\ie~KMR IIWA robot). 
We sample the robot base poses for arm robots such that all objects are reachable in both the start and the goal poses.
The same robot base poses were used for all the methods.
Next, we discuss the results for each task separately.

HPP~\cite{lamiraux2021hpp} fails to solve shelf tasks for more than one object in the given time limit.
Our method outperforms the PDDLStream planner~\cite{Symbolic_planning_Garrett_2020} based on the success rate.
For the KMR robot, all methods fail to solve the shelf task with three objects and the tunnel task, most likely due to the large base of the robot colliding with the environment.
To conclude, the main benefit of guiding the planner by video for the shelf task lies in the success rate, \ie~the guidance by video allows us to solve the given task more reliably.

For the tunnel task, the state-of-the-art planners fail to solve the task due to the difficulty of discovering the narrow passage by the sampling-based planning methods.
The proposed approach takes advantage of the available demonstration to discover the narrow passage and manages to plan a path for the non-mobile robots.

For the waiter task, we achive a higher success rate than the PDDLStream planner~\cite{Symbolic_planning_Garrett_2020}, however, it takes more time to arrive at a solution. The PDDLStream planner ignores the tray and moves each object one by one, which results in a lower number of grasps.

\subsection{Generalization}
\label{sec:experiment_generalization}
\begin{figure*}[t]
  \centering

    \includegraphics[width=\linewidth]{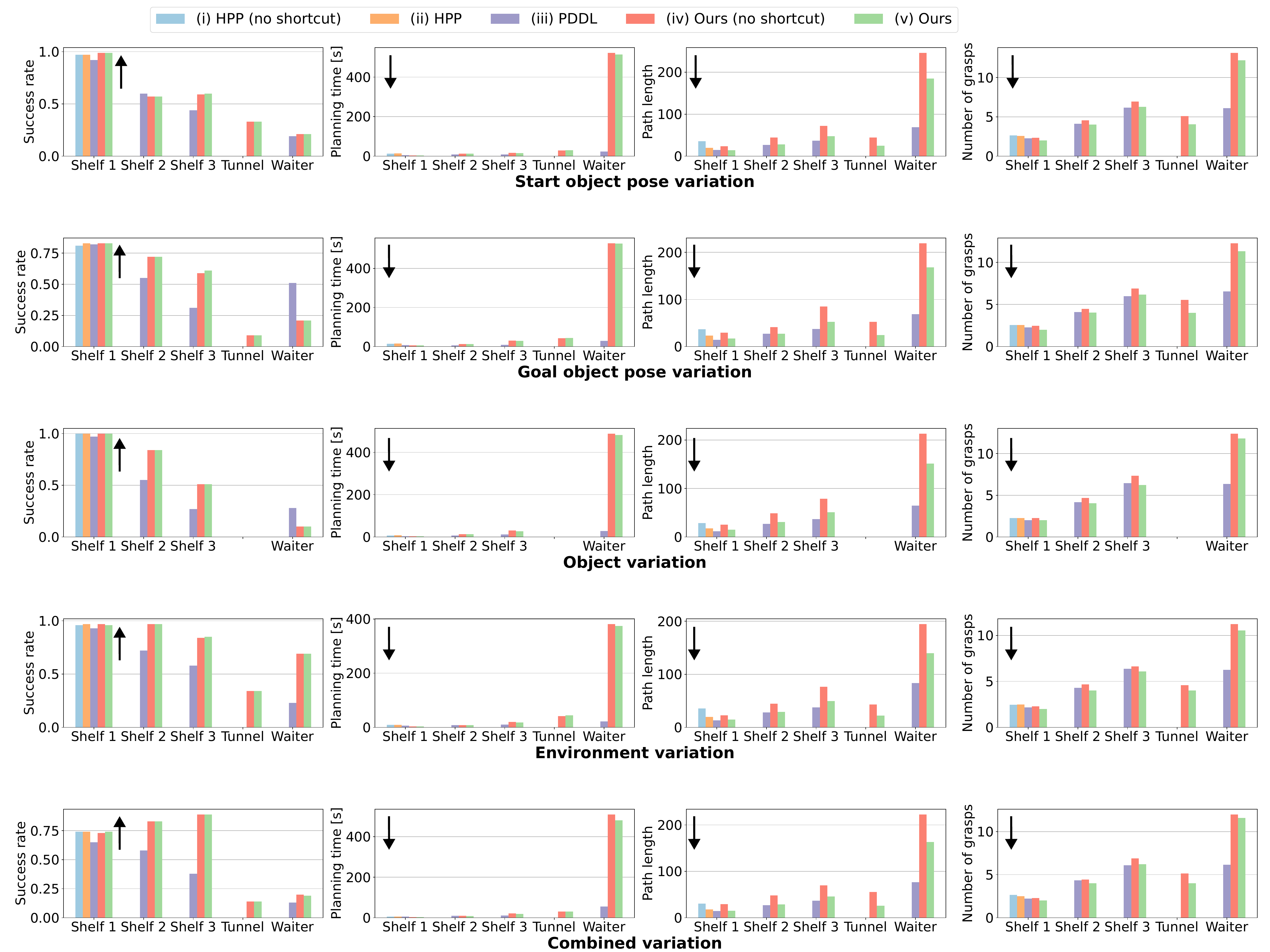}
  
  \caption[Generalization quantitative evaluation]{
  {\bf Quantitative evaluation of generalization capabilities.} 
  We study performance for the following generalization scenarios (see~\ref{sec:experiment_generalization} for more details): (i) start object pose variation: random modification of start object poses; 
  (ii) goal object pose variation: random modification of goal object poses; 
  (iii) object variation: replace original objects with randomly chosen novel objects;
  (iv) environment variation: random modification of the environment, \ie~furniture item pose;
  (v) combined variation: apply all aforementioned variations. 
  We report (from left): the success rate, the planning time [s], the path length, and the number of grasps.
  For the success rate, the higher number the better ($\uparrow$); for the other metrics lower numbers are better ($\downarrow$).
  The plots show the comparison of our method with the following baselines: (i)~RRT-connect implemented in \textcolor{hpp_tab}{HPP}~\cite{lamiraux2021hpp} without and (ii)~ \textcolor{hpp_opt_tab}{with the path optimization}, (iii)~\textcolor{pddl_tab}{PDDLStreams} planner~\cite{Symbolic_planning_Garrett_2020}, and (iv)~our \textcolor{our_tab}{proposed approach} without and (v)~\textcolor{our_opt_tab}{with the proposed path optimization}.}  
  \label{fig:generalization_quantitave}
\end{figure*}

We investigate the generalization capabilities of the proposed approach under the variation of: (i) the initial object pose, (ii) the goal object pose,  (iii) manipulated objects, (iv) the environment (the furniture pose), and (v) combined all aforementioned variations. The robot type and the robot pose are fixed. In Fig.~\ref{fig:generalization_quantitave} we report metrics described in~\ref{sec:experimetns_evaluation} averaged across 100 runs. 
For each run, variations of the respective variables were sampled randomly according to the distributions described next.
The parameters were resampled if all objects were not graspable by the same handle at the consecutive placement states. 

\begin{figure}
  \centering
    \includegraphics[width=0.48\textwidth]{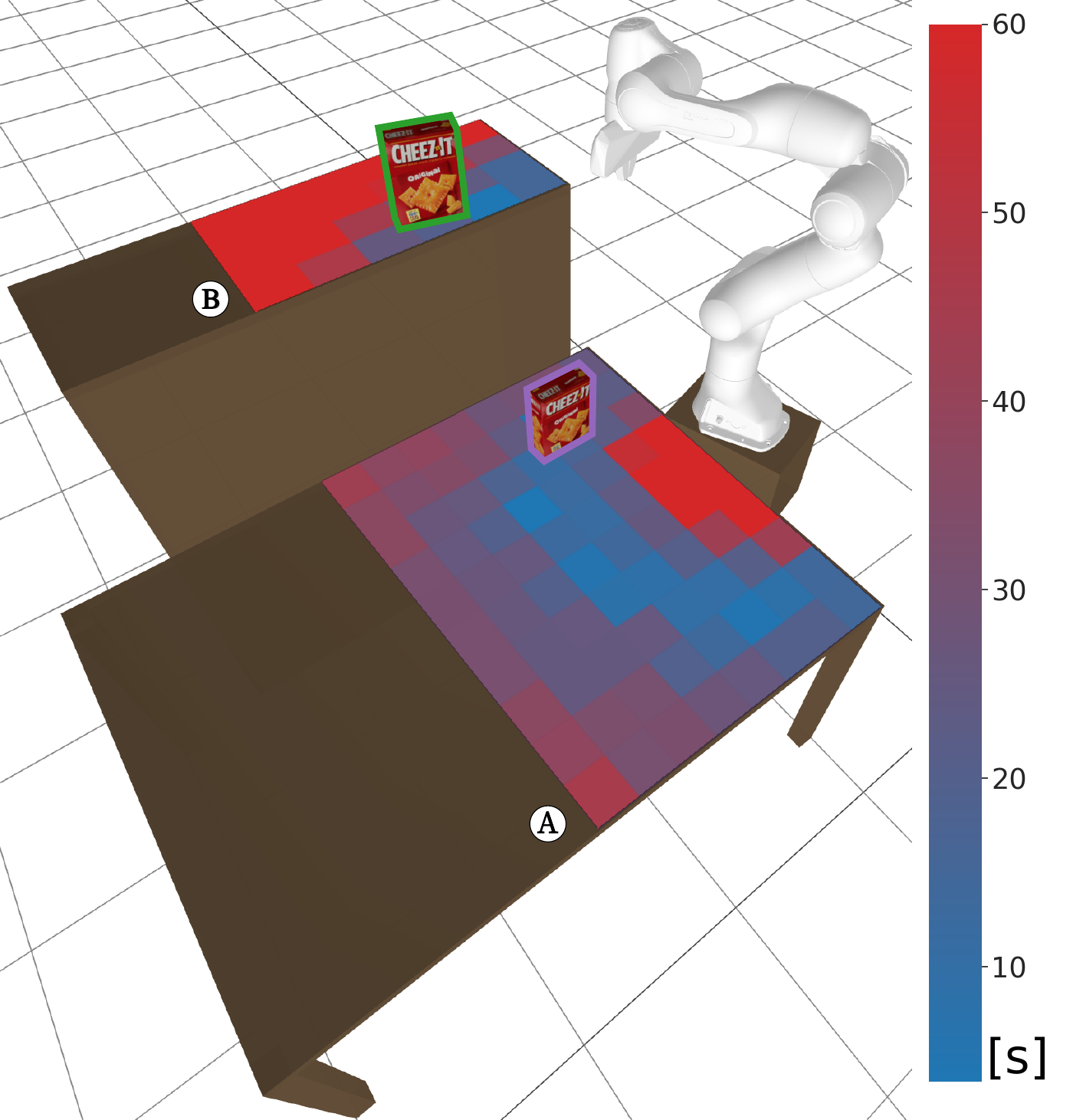}
  \caption[Visualization of generalization capabilities in 3D]{
  {\bf Planning time per object pose.} We show two heatmaps that show planning time (in seconds) when (A) the object start pose is varied, the goal pose is fixed on the shelf (highlighted with green), (B) the object goal pose is varied, the start pose is fixed on the table (highlighted with magenta). For every pose, we run our approach for 10 rotations around the object's $z$ axis and run the experiment with 5 different seeds per rotation. As evident from the plot, the planner takes longer~(or reaches the timeout) to find a valid path when the object is further from the robot's reach.
    }
  \label{fig:generalization_3d}
\end{figure}

\noindent \textbf{Start and goal object pose variation.} 
For the start pose modification, we fix the goal pose that is given by the demonstration and update the start pose (see Fig.~\ref{subfig:generalization_object_start}). We sample the $x$ and $y$ coordinate differences for the start pose from manually defined bounds and rotation around the $z$ axis, $\theta \in [-\pi; \pi]$. Bounds are chosen such that objects remain on the corresponding contact surface. The same process applies to the goal pose~(see Fig.~\ref{subfig:generalization_object_goal}).
Fig.~\ref{fig:generalization_quantitave} show quantitative results averaged across 100 runs.
In Fig.~\ref{fig:generalization_3d}, we report the average planning time per start and goal pose. The first two lines of  

\begin{figure*}
  \centering

    \includegraphics[width=\linewidth]{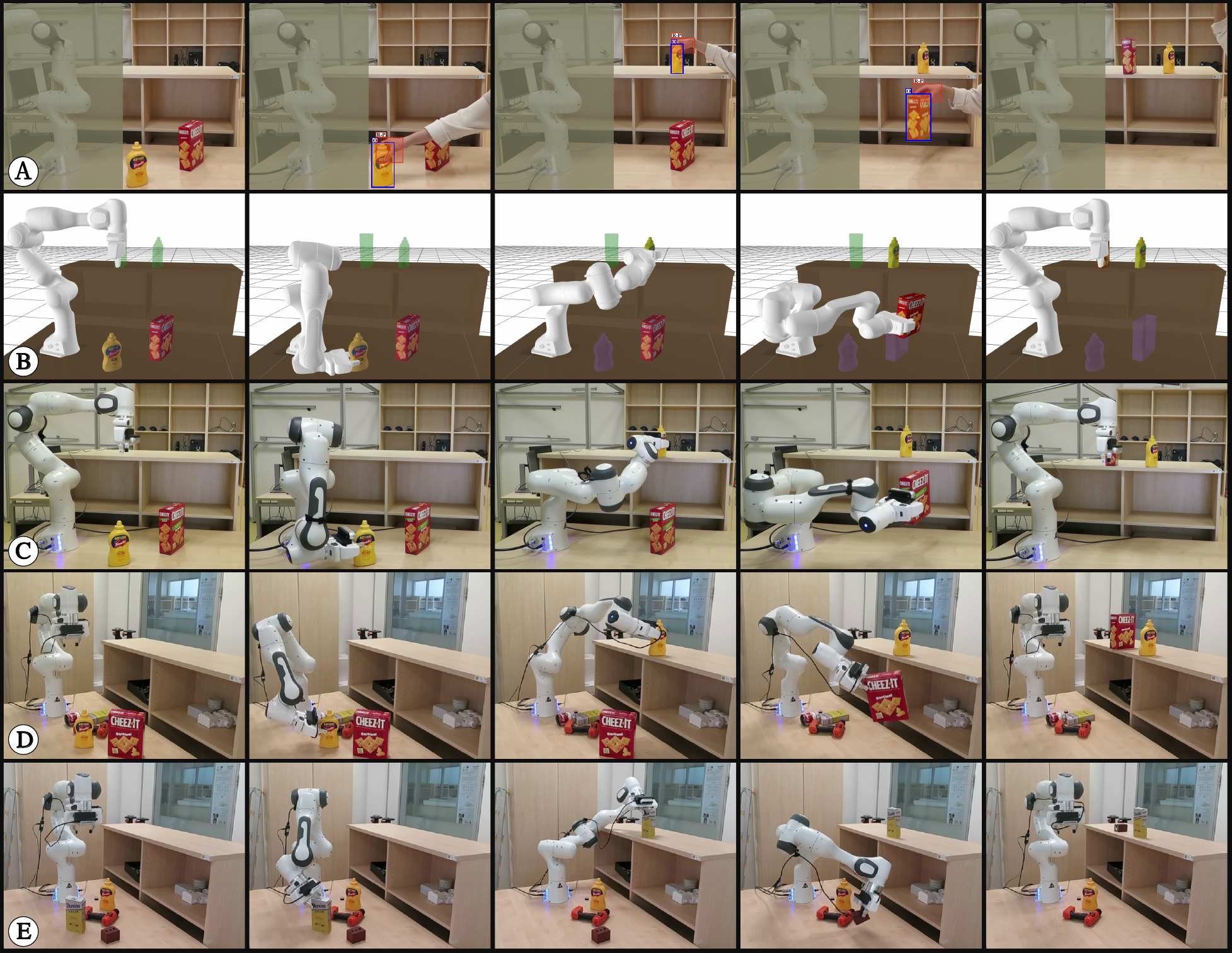}
  
  \caption[Real robot experiment.]{
  {\bf Real robot experiment.} The first line (A) shows the demonstration video with the recognized object pose and state of contact between the hand and the object (blue and red rectangles). The second line (B) shows the visualization of the planned trajectory. Here starting object poses are marked with violet color, and goal object poses are marked in green. The third line (C) shows the planned trajectory executed on the Franka Emika Panda robot. The fourth line (D) shows the output of trajectory refinement~(\ref{subsec:trajopt}) using the automatically estimated starting object poses~(as described in \ref{sec:experiment_generalization}). The fifth line (E) shows the output of trajectory refinement~(\ref{subsec:trajopt}) using the automatically estimated starting object poses of new objects~(the Domino sugar and the brown block) in changed  environment (shelf moved 20 cm). Both (D) and (E) use demonstration video (A) to guide the planning algorithm.
    }
      \label{fig:real_robot_exp}
\end{figure*}

\noindent \textbf{Object variation.} For each task, we sample a new set of objects out of available objects from the YCBV dataset (see~Fig.~\ref{subfig:all_possible_objects}). The start and the goal object poses are given by the demonstration. Since the tunnel task was designed to use a particular object, this scenario is not evaluated for the tunnel task. The third line of Fig.~\ref{fig:generalization_quantitave} shows quantitative results averaged across 100 runs. 

\noindent \textbf{Environment variation.} 
For each task, we sample $x$, $y$, and $z$ coordinate differences for the furniture piece from a uniform distribution (see Fig.~\ref{subfig:generalization_env}).
The bounds (in meters) for the shelf are $x \in [-0.5; 0.5]$, $y \in [0.0; 0.3]$, $z \in [-0.5; 0.5]$.  The bounds for the tunnel are $x \in [-0.1; 0.1]$, $y \in [-0.5; 0.1]$, $z \in [0.0; 0.0]$. The bounds for the table in the waiter task are $x \in [-1.0; 1.0]$, $y \in [-2.0; 0.0]$, $z \in [-0.2; 0.0]$.
The rotation of the furniture remains constant.
Poses of objects that rest on the surface of that furniture are changed accordingly.
The fourth line of Fig.~\ref{fig:generalization_quantitave} shows quantitative results averaged across 100 runs.

\noindent \textbf{Discussion of the generalization results}.
The trend is consistent with the quantitative results reported in Fig.~\ref{fig:restults}.
Only the shelf task with one object is solved by all methods.
Classical HPP~\cite{lamiraux2021hpp} fails to solve the other tasks in the given time limit.
Our method outperforms the PDDLStream planner~\cite{Symbolic_planning_Garrett_2020} based on success rate, but takes more time to arrive at a solution.
The tunnel task is only solved with our method since we take advantage of the video demonstration to find a narrow passage.
To conclude, the guidance from one video demonstration allows us to solve the given tasks for novel scenes without the need to shoot a new video demonstration.

\noindent \textbf{Real robot experiment.} The real robot experiment is shown in Fig.~\ref{fig:real_robot_exp}.
We fix the robot base pose on the table and plan based on the provided human demonstration.
The planned path is executed in the same environment with start and goal poses corresponding to the demonstration. 
We also demonstrate generalization capabilities for the real robot execution.  
For this, we take five images of the workspace by moving the camera attached to the robot around the workspace.
We use the consistent multiview multiobject pose estimation from CosyPose ~\cite{cosypose} to estimate the precise object pose measurements $\widehat{\mathcal{A}}_0$.
The object starting poses $ \mathcal{A}_0$ in the demonstration are replaced by $\widehat{\mathcal{A}}_0$, and these poses are used as described in the generalization section (Sec.~\ref{subsec:generalization}). 
We use our planner to find a feasible trajectory to complete the task with the updated object poses. The trajectory is further refined as described in Sec.~\ref{sec:path_optimization} and the resulting trajectory is executed on the robot (see Fig.~\ref{fig:real_robot_exp}, rows D and E). 
We additionally verify the benefits of the trajectory refinement and report the results in Table~\ref{tab:path_len}. We compare the running time and path length on the shelf task with 2 objects for all successful runs and we report the average numbers.
The results show that trajectory refinement can further decrease the length of the path by 33\% on average at the cost of additional computation burden (additional 60~s on average).

\begin{table}[!t]
\caption{
Trajectory refinement on the Shelf 2 task with the Panda robot.
Average computation time and path length are reported, showing the capability of the refinement algorithm to decrease the path length (at the cost of increased computation time). The path length is evaluated as the sum of Euclidean distances computed between the consecutive robot configurations in radians.
Note, that other metrics are not affected by the trajectory refinement and are therefore not reported here.}
\label{tab:path_len}

\centering
\begin{tabular}{lcc}
\toprule
 & Ours & Ours + trajectory refinement \\
\midrule
Computation time & 6 s & 67 s \\
Path length & 32.6 rad & 25.7 rad \\
\bottomrule
\end{tabular}
\end{table}

\section{Conclusion}
In this work, we propose a novel approach for guiding manipulation TAMP with video demonstration. From demonstrations, we extract object 6D poses and human-object contact changes. The extracted information is used to construct an admissible configuration space, which is efficiently explored by the RRT-like planner. The planner finds a feasible path that successfully solves the given task. This path is further refined by solving an optimal control problem to smooth robot motion.

\noindent\textbf{Limitations.} Our approach has several assumptions and limitations. We assume the demonstration consists of a sequence of pick-and-place motions (without pushing or sliding), where we manipulate one object at a time. 
The start and goal object poses should be reachable by the robot and suitable for grasping. For example, objects should not be too close to each other; otherwise, the planning will fail as the gripper will not be able to grasp the object.
For the waiter task, we assume that objects are manipulated only when the moving platform is stable.
Objects and hands should be well visible at the contact events for accurate processing of the video demonstration. Significant occlusions can still pose a challenge for the current approach.

\bibliographystyle{IEEEtran}
\bibliography{references}


\end{document}